\documentclass{article}

     \usepackage[final,nonatbib]{neurips_2018}

\usepackage[utf8]{inputenc} % allow utf-8 input
\usepackage[T1]{fontenc}    % use 8-bit T1 fonts
\usepackage{hyperref}       % hyperlinks
\usepackage{url}            % simple URL typesetting
\usepackage{booktabs}       % professional-quality tables
\usepackage{amsfonts}       % blackboard math symbols
\usepackage{nicefrac}       % compact symbols for 1/2, etc.
\usepackage{microtype}      % microtypography
\usepackage{booktabs}

\usepackage[colorinlistoftodos]{todonotes}
\usepackage{xspace}
\newcommand{\M}{\textsc{ML4H}\xspace}
\newcommand{\NP}{100\xspace}

\author{%
  Matthew B.A. McDermott\thanks{Equal Contribution} \\
  Massachusetts Institute of Technology\\
  \texttt{mmd@mit.edu} \\
  \And
  Shirly Wang$^*$ \\
  University of Toronto\\
  \texttt{shirlywang@cs.toronto.edu} \\
  \And
  Nikki Marinsek \\
  Evidation Health, Inc.\\
  \texttt{nmarinsek@evidation.com} \\
  \And
  Rajesh Ranganath \\
  New York University \\
  \texttt{rajeshr@cims.nyu.edu} \\
  \And
  Marzyeh Ghassemi \\
  University of Toronto, Vector Institute \\
  \texttt{marzyeh@cs.toronto.edu} \\
  \And
  Luca Foschini\\
  Evidation Health, Inc.\\
  \texttt{luca@evidation.com} \\
}

\title{Reproducibility in Machine Learning for Health}

\usepackage{todonotes}
\usepackage{framed}
\usepackage{enumitem}
\usepackage{multirow}

\begin{document}
\maketitle

\begin{abstract}
Machine learning algorithms designed to characterize, monitor, and intervene on human health (\M) are expected to perform safely and reliably when operating at scale, potentially outside strict human supervision. This requirement warrants a stricter attention to issues of reproducibility than other fields of machine learning.
%
% Therefore, due to potential far-reaching consequences, \M applications arguably demand higher guarantees of reproducibility than other fields in science.
In this work, we conduct a systematic evaluation of over \NP recently published \M research papers along several dimensions related to reproducibility. We find that the field of \M compares poorly to more established machine learning fields, particularly concerning data and code accessibility. Finally, drawing from success in other fields of science, we propose recommendations to data providers, academic publishers, and the \M research community in order to promote reproducible research moving forward.

% Reproducibility is a commonly touted goal in machine learning, yet as a field, we know that we often fall short.
% %
% In this work, we explore the goal of reproducibility, using analogies to the natural sciences to form a more general vision of reproducibility which captures our goal. Using this broad illustrate through argument and literature review that, under this broader definition, no sub-field of machine learning is this problem more paramount and more difficult than machine learning for health (ML4H). ML4H

% Reproducibility is a topic of great interest in the machine learning community, and great strides are being taken to ensure that models are repeatable, stable, and fully reproducible within the spectrum of their claims. However, in the subfield of Machine Learning for health.

% Health applications of ML are quickly gaining traction due to achieved performances. However, while the ML community seeks validation on observational datasets by making them open, the medical research community has historically favored replicability of findings in prospective, independent data collection.
% This has created a void in the reproducibility of ML in Health.
% In this work, we present a state of the art of reproducibility in ML health application. We note a void compared to other fields. We propose that:
% 1) benchmarks datasets should be proposed for health datasets and 2) a practice of generating simulated data should be adopted for the case data cannot be shared
\end{abstract}

\section{Introduction}
Science requires reproducibility, but many sub-fields of science have recently experienced a reproducibility crisis, eroding trust in processes and results and potentially influencing the rising rates of scientific retractions \cite{baker_1500_2016,cokol_retraction_2008,vasilevsky_reproducibility_2013}.
%In machine learning, the lack of reproducibility in papers reduces trust in machine learning as a whole.
Reproducibility is also critical for machine learning research, whose goal is to develop algorithms to reliably solve complex tasks at scale, with limited or no human supervision. Failure of a machine learning system to consistently replicate an intended behavior in a context different from which that behavior was defined may result in dramatic, even fatal, consequences~\cite{levin_tesla_2018}.
% but can be a challenge because of large number of choices available to practitioner for the use and evaluation of a method.
% Reproducibility is a core tenant of the natural sciences and a critical component of a successful machine learning project---without reproducibility,
%
Ranking prominently among machine learning applications that may put human lives at stake are those related to Machine Learning for Health (\M). In a field where applications are meant to directly affect human health, findings should undergo heavy scrutiny along the validation pipeline from research findings to applications deployed in the wild. For example, in 2018, 12 AI tools using \M algorithms to inform medical diagnosis and treatment were cleared by Food and Drug Administration (FDA) and will be marketed to and potentially used by millions of Americans~\cite{muoio_roundup:_2018}. Verifying the reproducibility of the claims put forward by the device manufacturer should thus be a main priority of regulatory bodies~\cite{parikh2019regulation}, extending the need for reproducible \M results beyond the machine learning research community.
%If these decisions were based on irreproducible results, then these tools will present a serious risk to the practice of care to real patients.
%In addition to enhancing safety, \M closely collaborates with the biomedical sciences, and poor reproducibility among ML studies results in a lack of confidence by our scientific colleagues \cite{ghosh_machine_2019}
%

%Though necessary in \M, several factors compound to make the reproducibility difficult.

%.
%
%Second, the healthcare application area places a high premium on accurate results---models that are inaccurate or fail to reproduce in deployment environments can hinder care and delay integral scientific progress. Last year, 12 AI tools for medical diagnosis were approved by the FDA/CDRH with no public peer review.

Unfortunately, several  factors relating to the availability, quality, and consistency of clinical or biomedical data make reproducibility especially challenging in \M applications.
In this work, we make several contributions. First, we present a taxonomy of reproducibility tailored to \M applications, and designed to capture reproducibility goals more broadly. Second, we use this taxonomy to define several metrics geared towards quantifying the particular challenges in reproducibility faced within ML4H, and conduct a comprehensive review of the published literature to support our claims and compare \M to machine learning more generally. Finally, we build on this analysis by exploring promising areas of further research for reproducibility in \M.

% We hope this work can help guide the conversation around reproducibility in machine learning for health going forward, by better grounding the discussion and codifying our shared goals, highlighting the current state of reproducibility in machine learning for health, and highlighting areas of opportunity that we deem particularly promising/necessary.\todo{Be more specific here}

\section{A Reproducibility Taxonomy}
The common understanding of reproducibility in machine learning can be summed up as follows:
%
%.
%
%\begin{leftbar}
A machine learning study is \textit{reproducible} if readers can fully replicate the exact results reported in the paper.
%\end{leftbar}
%
We will call this concept \textit{technical replicability}, as it is centrally concerned with whether or not one can replicate the precise, technical results of a paper under identical conditions. Though intuitive, we argue technical replicability is actually only a small part of the goal of ``reproducibility'' more generally. This discrepancy has been noted historically in various ways \cite{drummond_replicability_2009,gundersen_2018,joelle_pineau_reproducible_2018,plesser_reproducibility_2018} and is made apparent by the use of the term in the natural and social sciences, where attempted reproductions will often occur at different labs, using different equipment/staff, etc.
%
% Instead, more generally than technical replicability, reproducibility requires that the claims and results purported by a study are able to be reproduced to the fullest extent of their conceptual limits. This distinction is most critical for ML4H, not least because this notion is much closer to that used in the biomedical sciences
%
We argue that in order for a study to be fully reproducible, it must meet three tiers of replicability:

\SetLabelAlign{parright}{\parbox[t]{\labelwidth}{\raggedleft#1}}
\begin{description}[
leftmargin=!,labelwidth=\widthof{\bfseries Conceptual Replicability},align=parright
]
    \item[Technical Replicability] Can results be replicated under \emph{technically} identical conditions?
    \item[Statistical Replicability] Can results be replicated under \emph{statistically} identical conditions?
    \item[Conceptual Replicability] Can results be replicated under \emph{conceptually} identical conditions?%\footnote{Note that this notion is task/use-case specific. If a paper purports its result holds in only a very limited context, then the burden here is smaller than if broader claims are made.}
\end{description}

\emph{Technical replicability} refers to the ability of a result to be fully technically replicated, yielding the precise results reported in the paper. For example, this entails aspects of reproducibility related to code and dataset release. \emph{Statistical replicability} refers to the ability of a result to hold under re-sampled conditions that yield different technical configurations, but should not statistically affect the claimed result (e.g., a different set of random seeds, or train/test splits). Note this is related, but not identical, to the notion of \emph{internal validity}~\cite{campbell1986relabeling} commonly used in social science research. Similarly, \emph{conceptual replicability} is heavily related to \emph{external validity}~\cite{pearl2014external}, as it describes the notion of how well the desired results reproduces under conditions that mach the conceptual description of the purported effect. Note that this is task-definition dependent; claiming a task has a greater conceptual horizon of generalizability makes it harder to satisfy this reproducibility requirement.

All three of these replicability criteria are central for full reproducibility: without technical replicability, one's result cannot be demonstrated. Without statistical replicability, one's result will not reproduce under increased sampling and the presence of real-world variance. And lastly, without conceptual replicability, one's result does not depend on the desired properties of the data, but instead depends on potentially unobserved aspects of the data generation mechanism that, critically, \emph{will not reproduce when deployed in practice}.

Under each of these lenses, ML4H differs from general machine learning domains in critical ways and presents unique challenges.

\section{Core Reproducibility Challenges in ML4H}
\begin{figure}[t]
    \centering
    \includegraphics[width=1\textwidth]{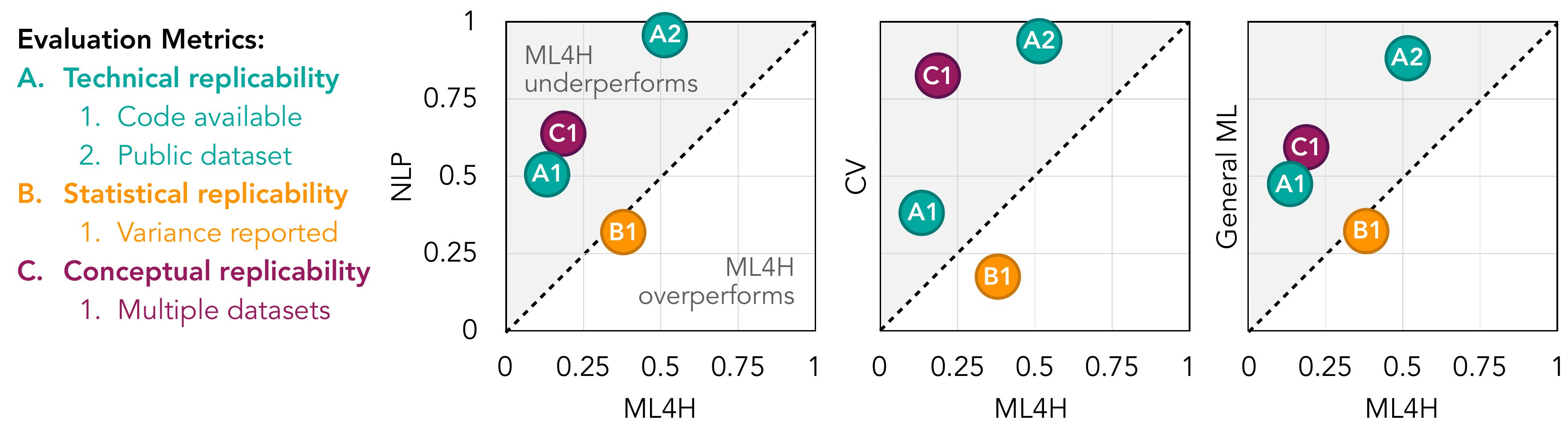}
    \caption{Fraction of papers satisfying certain conditions by ML field. See the Appendix (Section~\ref{sec:review_procedure}) for detailed descriptions of the underlying data collection procedure. Note that ML4H consistently lags other subfields of machine learning on all measures of reproducibility save inclusion of proper statistical variance.
    }
    \label{fig:paper_stats}
\end{figure}

In this section, we illustrate both through qualitative arguments and a quantitative literature review that ML4H lags behind other subfields of machine learning on various reproducibility metrics. Our literature review procedure entailed manually extracting and annotating over 300 papers from various venues, spanning ML4H, Natural Language Processing (NLP), Computer Vision (CV), and general machine learning (general ML).\footnote{NLP and CV were chosen to represent broad fields with a significant applied focus, much like ML4H. General ML was chosen to have an unbiased comparison to the field more broadly.} The full procedure used for this statistical review is detailed in the Appendix (Section~\ref{sec:review_procedure}), and final quantitative results on several key metrics can be seen in Figure~\ref{fig:paper_stats}, though they are also detailed in the text where appropriate. The full set of results is available at https://github.com/evidation-health/ICLR2019.

\subsection{Technical Replicability}
ML4H faces several key challenges with regards to technical replicability. Firstly, health data is privacy sensitive, making it difficult to release openly without either incurring risks of re-identification, or severely diminishing utility by applying aggressive de-identification techniques~\cite{dwork_ullman_2018}. As a result, few public datasets are available, and those that are available are used extremely frequently, leading to a risk of dataset-specific over-fitting. To this point, approximately only 51\% of the ML4H papers we examined used public datasets, as compared to over 90\% of both CV and NLP papers, and approximately 77\% of general ML papers.

ML4H scores even more poorly when it comes to code release, preprocessing specification, and cohort description; only approximately 13\% of the papers we analyzed released their code publicly, compared to approximately 37\% in CV and 50\% in NLP. A previous study that reviewed the prevalence of code release in AI reported that only 6\% of papers released code \cite{gundersen_2018}, which is lower than all of our estimates. This discrepancy may be due to sampling papers from different conferences and different time periods (2013-2016 vs. 2018 for the current study). Specific prior works have also already examined the prevalence of code release and dataset cohort sub-selection within ML4H, finding that even when restricting focus to public datasets, studies often do not release code or include sufficiently informative text to enable a full technical reproduction \cite{hegselmann_reproducible_2018,johnson2017reproducibility}.  Note that code release is itself not necessarily sufficient for full technical replicability; even when code is released, it may not run correctly, it may exclude critical details, or it may fail to generate the results reported in the paper.

% Finally, within what data we do have and release, benchmarks are uncommon; instead, researchers often define their own cohorts, preprocessing methods, and tasks to demonstrate their models, leading to a dearth of comparable published models~\cite{harutyunyan2017multitask}.

\subsection{Statistical Replicability}
To assess the state of statistical replicability in ML4H, we quantified how often papers describe the variance around their results (e.g., by listing both the mean and the standard deviation of a performance metric over several random splits). Interestingly, while ML4H's rate of this is still relatively low (approximately 38\%), it is higher than that of CV, NLP, or the general domain (17\%, 33\%, and 32\%, respectively).

Though this is an encouraging sign, we feel there is still significant room for improvement.
% Healthcare data is noisy, and we often have relatively small, high-dimensional data to work with, rendering overfitting more likely, so we must be vigilant in our best practices to ensure we produce statistically viable results.
%
Even in other fields of machine learning, with arguably less complex data types, repeated studies have shown that published models fail to statistically reproduce when appropriate statistical procedures are implemented and fair hyperparameter search/preprocessing methods are used: in \cite{recht_imagenet_2018}, researchers find that published models evaluated using the public ImageNet test set show consistent performance drops when trained/tested on other random splits within ImageNet, and in \cite{melis_state_2017,lucic_are_2017}, researchers find that published performance deltas between various model architectures for language models and generative adversarial networks, respectively, fail to persist under more robust hyperparameter search and statistical comparison techniques. In \cite{henderson_deep_2017}, researchers examine practices in deep reinforcement learning which can limit or enhance studies' reproducibility. Due to our inability to technically replicate many ML4H works given dataset and code release issues, we cannot assess the extent of these issues to the same degree in ML4H. However, it seems doubtless that similar problems affect our field, perhaps to an even greater degree given that our datasets are frequently relatively small, high dimensional, sparse/irregularly sampled, and suffer from high rates of noise.

\subsection{Conceptual Replicability}
The critical issue in ML4H preventing conceptual replicability is the lack of multi-institution datasets in healthcare and the lack of usage of those that do exist. Whereas approximately 83\% of CV studies and approximately 66\% of NLP studies used multiple datasets to establish their results, only approximately 19\% of ML4H studies did. Using only a single dataset is potentially dangerous, as it is known that developing ML4H models that generalize over changing care practices or data formats is challenging. In \cite{gong_predicting_2017}, researchers demonstrate the dangers of training a model on raw data from one institution and transferring to another (using a simulated institutional divide within the MIMIC dataset), and in \cite{nestor_rethinking_2018}, researchers establish that without using manually engineered representations, ML4H models exhibit significant degradation in performance over time as care patterns evolve. These results are not surprising; health data is rife with hidden confounders, differs significantly between data collection and deployment environments, drifts over time, and further differs in structure and concept between different healthcare institutions~\cite{caruana2015intelligible}.

\section{Opportunities for Improvement}
\begin{figure}
    \centering
    \includegraphics[width=0.6\columnwidth]{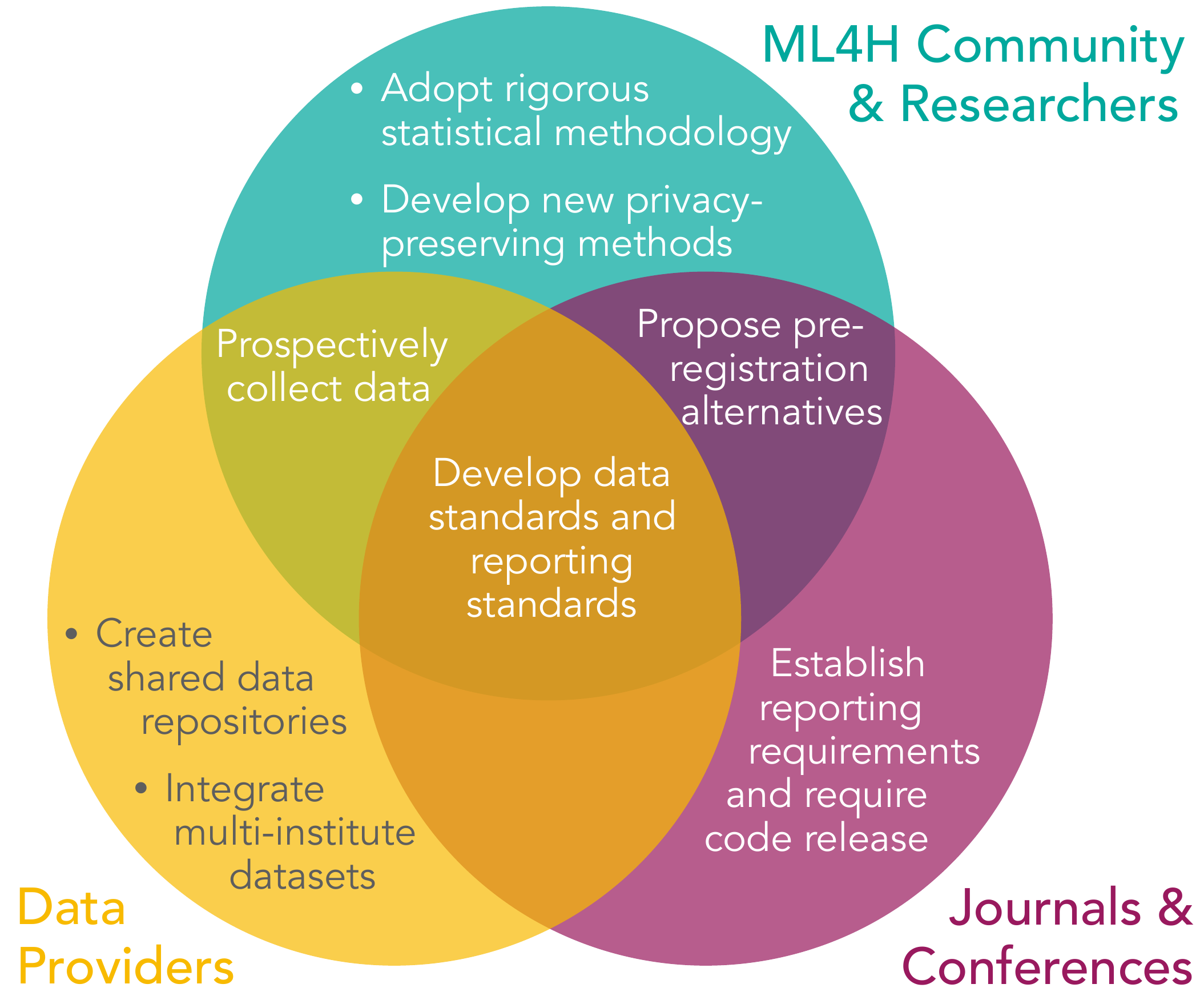}
    \caption{Summary of recommendations for different stakeholders.
    }
    \label{fig:recommend}
\end{figure}

In this section, we present several practical suggestions for enhancing reproducibiltiy for ML4H from the perspective of three main \M stakeholders: the \M research community, data providers, and related journals and conferences.  Figure~\ref{fig:recommend} summarizes these suggestions.

\subsection{Create Shared Research Resources}
%As machine learning becomes profitable with applications in vision, speech, natural language processing, and health, it is important that researchers create and maintain public sources of data and code that can maintain cutting-edge technology. Many large technology corporations are making strong moves towards purchasing large amounts of clinical data \cite{wood_improving_2019,rajkomar_scalable_2018}. While such efforts may create resources for sale to hospitals, recent patent filings \cite{bresnick_google_2019,google_inc._system_nodate} also provide cautionary notes for the community. With the creation of large data trusts where medical institutions can anonymously pool data for researchers to use and create algorithms, the medical and machine learning community can continue to leverage resources and maintain an aggressive pace of research. Such a collective could also allow for shared instance hosting, where community data queries and code are released for verification internally to a core group of trusted users
%(cite ML Perf community benchmarks here)
%before being released more broadly.
Data providers such as hospitals, clinical research centers, and government agencies produce vast amounts of valuable data. Unfortunately, as suggested by our literature review, few datasets are available for a wide range of researchers to explore. We call for more instances of large data trusts where medical institutions can anonymously pool data for researchers to use and create algorithms. Several prominent examples already exist for the field to draw on in creating such repositories, including MIMIC~\cite{johnson2016mimic}, the U.K and Japan Biobanks~\cite{nagai2017overview,sudlow2015uk}, eICU~\cite{pollard_eicu_2018}, the Temple University Hospital EEG Corpus~\cite{harati2014tuh}, and Physionet~\cite{goldberger2000physiobank}. This recommendation is especially compelling as corporate entities increasingly invest in the ML4H space, particularly with regards to data~\cite{rajkomar_scalable_2018,wood_improving_2019}. Anchoring our research progress to large scale datasets in non-public or non-academic hands poses a dangerous precedent, especially in light of recent patent filings~\cite{bresnick_google_2019,google_inc._system_nodate}.
%\todo{cite datasets}

\subsection{Integrate Multi-Institute Datasets}
%A particular focus area for subsequent data releases should be the inclusion of additional \emph{multi-institution} datasets.
Multi-institute datasets enable studies to assess their ability to translate to new contexts, a critically understudied facet of ML4H research. Recent strides have been taken in this domain with the release of the eICU dataset~\cite{pollard_eicu_2018}, one of ML4H's first large scale, multi-institution EHR datasets, and researchers are already analyzing how to form generalizable models using this resource \cite{johnson_generalizability_2018}. Lastly, Observational Health Data Sciences and Informatics (OHDSI)~\cite{hripcsak2015observational} provides a mechanism to run observational health studies across multiple institutions and countries.
We encourage more collaborative efforts in this area from data providers to improve conceptual replicability.

\subsection{Prospectively Collect Data}
Data collected as a by-product of care, then later released for research purposes (e.g., MIMIC \cite{johnson_mimic-iii_2016}), presents serious privacy risks and contains many confounding variables. The landscape of these privacy risks and the nature of the confounding variables change (though they do not necessarily lessen) if data is instead prospectively collected directly from new, consented participants. These types of data collection regimes are logistically challenging, but are possible.
%
% ML4H researchers can work with data providers on alternative data collection methodologies which involve directly consenting participants are also feasible, though they have significant logistical burdens.
The NIH's All of Us Research Program \cite{national_institutes_of_health_all_nodate}, Evidation's DiSCover Project \cite{evidation_health_digital_2018}, and Google's Project Baseline \cite{verily_why_2017}
%, and Evidation's DiSCover Project~\cite{discover2018}
are examples of these approaches.

\subsection{Adopt Rigorous Statistical Methods}
ML4H researchers should be more rigorous in the development, refinement, and dissemination of statistical best practices (e.g., the proper procedures for model comparisons, etc.). Holding ourselves to high standards of statistical rigor, potentially including periodic statistical audits of our own statistical replicability, will help ensure we mitigate the problems other fields have found with ``over-fitting via publication,'' e.g. \cite{recht_imagenet_2018}.  Several commonly-used challenges already exist in ML4H with fixed train/test sets (e.g., \cite{ghassemi_you_2018,stubbs_annotating_2015}) which could be used for these kind of post-hoc replication studies.

\subsection{Develop New Privacy-Preserving Analysis Techniques}
Technological solutions can also be employed to help mitigate privacy concerns. ML4H researchers can explore noised, fully- or partially- simulated, or encrypted datasets. In cases where data cannot be released, techniques to train distributed models without sharing data have been proposed~\cite{vepakomma2018split}. Synthetic data can be an excellent tool to help enable researchers to still meaningfully release their code with full end-to-end realizations of their pipeline, as is done in, e.g., \cite{zink_fair_2019}. Technology for producing synthetic patient-level data also already exists \cite{walonoski_synthea:_2018}.

\subsection{Propose Pre-Registration Alternatives}
%Even beyond these difficulties, ML4H's close ties to the biomedical sciences necessitate a more rigorous commitment to statistical replicability.
In the biomedical sciences, ``observational studies,'' under which definition nearly all ML4H research falls, undergo intense scrutiny to ensure they are not susceptible to statistical artifacts---in particular, increasingly these studies are required to be \emph{pre-registered},\footnote{
Meaning they are required to report their goal and planned preprocessing/modeling scheme before they run any experiments in order to avoid intentional or unintentional statistical fraud. This idea is not without its detractors \cite{lash_commentary:_2012,editors_registration_2010}.
} a move promoted by scientists and major journals alike \cite{lancet_should_2010,loder_registration_2010,williams_registration_2010}. Such prospective checks make unintentional statistical fraud more difficult, but are utterly absent in ML4H. A verbatim application of the pre-registration practice in use in the epidemiology community is likely an impractical solution due to the intrinsic exploratory nature of model development, but other techniques, such as systematic re-release of new data and/or rotation of official train/test splits have been found to help reduce the presence of statistical errors in other fields. ML4H researchers and academic publishers should engage in serious conversation on what the best vehicle for these checks and balances are.
%For interventional studies (clinical trials, in particular) this practice has had a dramatic impact, dropping the rates of positive results found in clinical trials by significant magnitudes \cite{kaplan_likelihood_2015}. This practice of pre-registering, or even more generally fully planning an analysis pipeline prior to performing significant data exploration is utterly absent in ML4H. This should make us rightly wary of our statistical replicability.\\

%In traditional biomedical sciences, statistical auditing techniques such as pre-registration, designed to make unintentional statistical fraud more difficult, have had a dramatic impact. Are there strategies we can deploy in ML4H in this vein to have a similar effect? Pre-registration is likely too aggressive a solution at this early stage for the field, but other techniques, such as systematic re-release of new data and/or rotation of official train/test splits have been found to help reduce the presence of statistical errors in other fields and could serve us well here.

% \paragraph{Self-Monitoring}
% No matter how vigilant we are internally in our use of statistical best practices, we will not know how consistent our result streams are until we measure it. Similar to works taken in other fields assessing published model's statistical replicability (e.g., \cite{recht_imagenet_2018}), we must also assess our models regularly.

\subsection{Establish Reporting Requirements and Require Code Release}
If conferences and journals required greater rates of data/code release, or the use of additional data/code availability statements at publication, this would instill a significant pressure on the field to address some of the foundational barriers to reproducibility. Conferences and journals have the additional ability to insist on high standards for the reporting of statistical variance around results, hyperparameter search procedures, and evaluation mechanisms, each of which would help ensure that we maintain high standards of technical and statistical replicability.

\subsection{Develop Data Standards and Reporting Standards}
%Data in healthcare also lacks a standard format. Standards do exist, such as the OMOP standard \cite{overhage_validation_2012} or the FHIR standard \todo{cite}, but they are not commonly used in \M research.
Collaborative efforts in developing data standards and reporting standards is another avenue for improving reproducibility. Healthcare analytics organizations have developed data standards like the OMOP standard \cite{overhage_validation_2012} or the FHIR standard \cite{hl7_introducing_2018}, but they are not commonly adopted in \M research. Increased use of standards would make it easier to technically and conceptually replicate \M studies. Similarly, when \M datasets are created, we need greater descriptions of their contents, potential confounders and biases, missing data prevalence and distribution, and how they were created. Increasing the use of ``specs'' or ``datasheets'' describing datasets would help allay these concerns---efforts in the broader machine learning community are also approaching this goal \cite{gebru_datasheets_2018} and we should vigorously adopt these practices in \M. %In ML4H specifically, additional concerns may need to be noted in the datasheets, such as the levels and types of observed data missingness.

%\subsection{Statistical Analyses}
%How our field performs statistical analyses also warrants increased attention in several axes.

\section{Conclusion}
In this work, we have framed the question of reproducibility in \M around three foundational lenses: technical, statistical, and conceptual replicability. In each of these areas, we argue both qualitatively and quantitatively, through a manual, extensive review of the literature, that \M performs worse than other machine learning fields in several reproducibility metrics we have identified. While keeping in mind the intrinsic challenges of data acquisition and use that plague the field, we highlight several areas of opportunities for the future, focused around improving access to data, expanding our trajectory of statistical rigor, and increasing the use of multi-source data to better enable conceptual reproducibility.

\section*{Acknowledgments}
This paper benefited substantially from the help of many people. Most notably, Bret Nestor, Amy Lu, Denny Wu, Elena Sergevea, and Di Jin all helped annotate papers for our analysis.

Additionally, this work was funded in part by National Institutes of Health: National Institutes of Mental Health grant P50-MH106933, a Mitacs Globalink Research Award, as well as a CIFAR AI Chair at the Vector Institute, and an NSERC Discovery Grant.

\bibliography{iclr2019_conference,matthew_bibliography}
\bibliographystyle{plain}

\section*{Appendix: Statistical Review Procedures}
\label{sec:review_procedure}

\paragraph{Selection Criteria}
Papers were selected at random, to ensure an unbiased sample, from various venues associated with different domains (though papers were tagged with their content-driven domain at annotation time). A full list of venues, along with the number of papers analyzed from each, is presented in Table~\ref{tab:paper_sources}. All publicly-accessible papers within the \M venues were used, rather than a random sub-sample, given the venues' limited sizes. Except for papers from MLHC2017 and ICCV, all papers were published in 2018. We conducted a similar review for 2018 papers only and the results were consistent.

\begin{table}[h]
    \centering
    \begin{tabular}{llr} \toprule
        Domain             & Conference               & \# Papers Examined \\ \midrule
        \multirow{4}{*}{ML4H} & MLHC 2017                & 25$^*$   \\
                              & MLHC 2018                & 31$^*$   \\
                              & ML4H Poster-Accepts 2018 & 57$^*$   \\
                              & KDD Health Day 2018      & 21$^*$   \\ \midrule
        \multirow{3}{*}{NLP}  & ACL                      & 30 \\
                              & EMNLP                    & 30 \\
                              & NAACL                    & 30 \\ \midrule
        \multirow{3}{*}{CV}   & CVPR                     & 21 \\
                              & ICCV                     & 21 \\
                              & ECCV                     & 21 \\
        \midrule
        \multirow{5}{*}{ML}   & NeurIPS                  & 14 \\
                              & ICML                     & 14 \\
                              & ICLR                     & 14 \\
                              & KDD                      & 14 \\
                              & AAAI                     & 14 \\
        \bottomrule
    \end{tabular}
    \caption{
        Sources and coverage statistics for our manual literature review. $^*$ indicates all publicly-accessible papers published were used.
    }
    \label{tab:paper_sources}
\end{table}

\paragraph{Annotation Procedure}
Each paper was reviewed by only one annotator, and in total 8 annotators were used, 3 of which are authors on this work. Annotation tasks were designed to be quick, to enable us to profile a large number of papers at a coarse level rather than a small number of papers in depth. As such, annotators were only asked the following questions:
\begin{enumerate}
    \item Was code released?
    \item What datasets were used?
    \item Are these datasets publicly available (modulo data use agreements)?
    % \item Do these datasets contain any PHI
    \item Do the authors report any notion of variance around their results or assess their comparisons to baselines in a statistically robust fashion (e.g., via hypothesis testing)?
\end{enumerate}

\paragraph{Potential Biases}
This selection and annotation procedure allowed us to analyze a large number of papers, but has several possible biases. In particular, our annotation questions were all of these were designed to be determinable via quick, scanning techniques and as a result this task took on average between 45 seconds and 3 minutes per paper. In such a limited time, some losses are unavoidable. We recognize several sources of possible bias worth mentioning.

Firstly, some papers may, for example, release datasets or code products external to the paper and not mention it in the actual text. We will omit these associated products. If such effects induce a notable bias in our results, however, we must question why as a field we are comfortable releasing our code/data without any mention in the associated paper.

Secondly, not all papers intended to be analyzed were publicly accessible. Similarly, the versions of papers we analyzed could have been different from the version presented at the actual conference venue, or there could exist updated versions of papers we analyzed in different repositories. Our analysis technique will miss these effects.

Thirdly, some papers naturally fit into multiple categories (e.g., a work focused on medical named entity recognition would be both a \M work and an NLP work). In the interest of ensuring our comparison classes were as pure as possible, we omitted all clearly multi-domain works, but allowed works that centered primarily in a single domain to remain.

Lastly, different fields present different kinds of works, and not all works fit into our framework. Largely theoretical works, for example, often have no real datasets or public experiments. Similarly, presenting variance is a different question for works focused principally around computational efficiency rather than predictive accuracy. We handled these issues by attempting to answer these questions as best we could, and flagging any papers that overtly did not fit our scheme and excluding them from our analyses.

\paragraph{Data Release}
We release all data as a result of these analyses publicly, accessible at https://github.com/evidation-health/ICLR2019.

\end{document}